\documentclass[10pt,twocolumn,letterpaper]{article}

\usepackage[pagenumbers]{cvpr} 
\newcommand{\fixed}[1]{\textcolor{black}{#1}}
\usepackage{multirow}
\usepackage{booktabs}

\usepackage[dvipsnames]{xcolor}

\definecolor{cvprblue}{rgb}{0.21,0.49,0.74}
\usepackage[pagebackref,breaklinks,colorlinks,citecolor=cvprblue]{hyperref}
\usepackage{algorithm}
\usepackage{algpseudocode}

\title{Representative Feature Extraction During Diffusion Process \\for Sketch Extraction with One Example}

\author{Kwan Yun$^*$
\hspace{4mm}
Youngseo Kim$^*$
\hspace{4mm}
Kwanggyoon Seo
\hspace{4mm}
Chang Wook Seo
\hspace{4mm}
Junyong Noh \vspace{3mm}\\
KAIST, Visual Media Lab\\
}

\begin{document}

\twocolumn[{
\maketitle
\begin{center}
    \vspace{-1.0em}
    \begin{tabular}{c}
        \hspace{-5mm}
        \includegraphics[width=\linewidth]{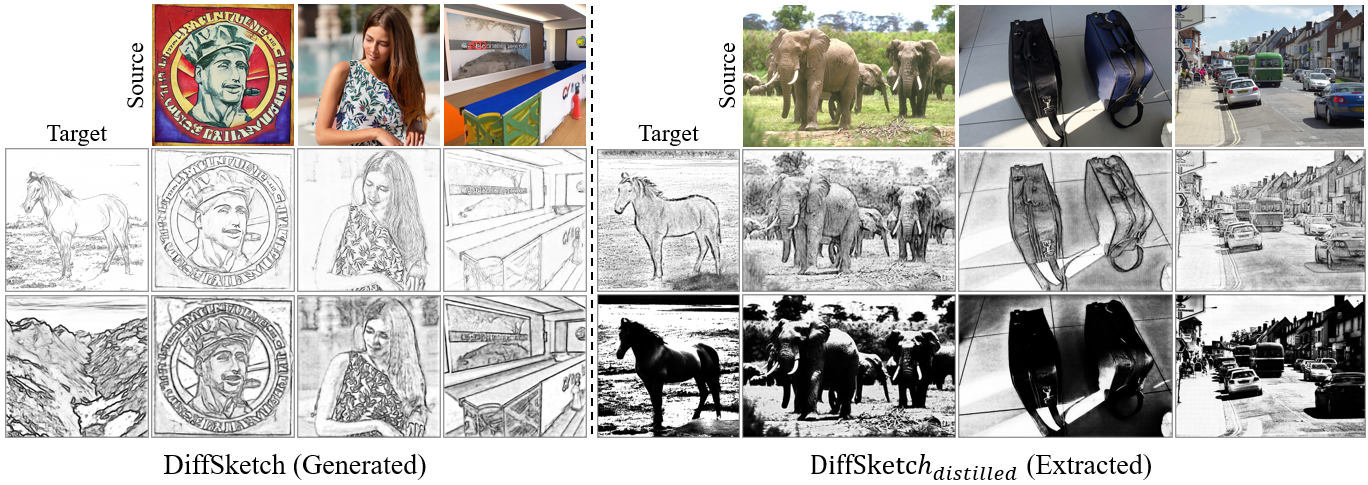}
    \end{tabular}\vspace{-1.0em}
    \captionof{figure}{{Results of DiffSketch and distilled $\text{DiffSketch}_{distilled}$, trained with one example. The left sketches were generated by DiffSketch, while the right sketches were extracted from images using $\text{DiffSketch}_{distilled}$.}}
\end{center}}]

\def\thefootnote{*}\footnotetext{These authors contributed equally to this work}\def\thefootnote{\arabic{footnote}}

\begin{abstract}
\vspace{-2mm}
We introduce DiffSketch, a method for generating a variety of stylized sketches from images. Our approach focuses on selecting representative features from the rich semantics of deep features within a pretrained diffusion model. This novel sketch generation method can be trained with one manual drawing. Furthermore, efficient sketch extraction is ensured by distilling a trained generator into a streamlined extractor. We select denoising diffusion features through analysis and integrate these selected features with VAE features to produce sketches. Additionally, we propose a sampling scheme for training models using a conditional generative approach. Through a series of comparisons, we verify that distilled DiffSketch not only outperforms existing state-of-the-art sketch extraction methods but also surpasses diffusion-based stylization methods in the task of extracting sketches.
\vspace{-2mm}
\end{abstract}   
\vspace{-1.0em}
\section{Introduction}
\label{sec:intro}

Sketching is performed in the initial stage of artistic creation of drawing, serving as a foundational process for both conceptualizing and conveying artistic intentions. It also serves as a preliminary representation that visualizes the core structure and content of the eventual artwork. As sketches can exhibit distinct styles despite their basic form composed of simple lines, many studies in computer vision and graphics have attempted to train models for automatically extracting stylized sketches \cite{winnemoller2011xdog,lllyasviel2017sketchKeras,ashtari2022reference,chan2022learning, seo2023semi} that differ from abstract lines~\cite{mo2021general,vinker2022clipasso,willett2023stylizing}.

Majority of current sketch extraction approaches utilize image-to-image translation techniques to produce high-quality results. These approaches typically require a large dataset when training an image translation model from scratch, making it hard to personalize the sketch auto-colorization~\cite{ci2018user,kim2019tag2pix,zhang2018two,yuan2021line} or sketch-based editting~\cite{liu2022deepfacevideoediting,zhang2023adding,seo2022midms,portenier2018faceshop}. On the other hand, recent research has explored the utilization of diffusion model \cite{rombach2022high,saharia2022photorealistic} features for downstream tasks \cite{xu2023open, khani2023slime, zhang2023tale, tumanyan2023plug}. Features derived from pretrained diffusion models are known to contain rich semantics and spatial information \cite{tumanyan2023plug, xu2023open}, which is known to help the training with limited data \cite{baranchuk2021label}. Previous studies have utilized these features extracted from a subset of layers, certain timesteps, or every specific intervals. Unfortunately, these hand-selected features often do not contain most of the information generated during the entire diffusion process.

To this end, we propose Diffsketch, a new method that can extract representative features from a pretrained diffusion model and train the sketch extraction model with one data. For feature extraction from the denoising process, we statistically analyze the features and select those that can represent the whole feature information from the denoising process. Our new generator aggregates the features from multiple timesteps, fuses them with VAE features, and decodes these fused features.

The way we train the generator with synthetic features differs from that employed by previous diffusion-based stylization methods in that our method is specially designed for sketch extraction. While most diffusion-based stylization methods adopt the original pretrained diffusion model by swapping features \cite{hertz2022prompt,tumanyan2023plug} or by inverting style into a certain prompt \cite{gal2023textualinversion,ruiz2023dreambooth}, these techniques do not provide fine control over the style of the sketch, making them unsuitable for extracting sketches in a desired style. In contrast, DiffSketch trains a generator model from scratch specifically for sketch extraction of a desired style.

In addition to the newly proposed model architecture, we introduce a method for effective sampling during training. It is easy to train a network with data that share similar semantic information to ground truth data. However, relying solely on such data for training will hinder the full utilization of the capacity provided by the diffusion model. Therefore, we adopt a new sampling method to ensure training with diverse examples while enabling effective training. Finally, we distill our network into a streamlined image-to-image translation network for improved inference speed and efficient memory usage. The resulting $\text{DiffSketch}_{distilled}$ is the final network that is capable of performing a sketch extraction task. The contributions can be summarized as follows:
\begin{itemize} 

\item We propose DiffSketch, a novel method that utilizes features from a pretrained diffusion model to generate sketches, learning from one manual sketch data.

\item Through analysis, we select the representative features during the diffusion process and utilize the VAE features as fine detailed input to the sketch generator.

\item We propose a new sampling method to train the model effectively with synthetic data.

\end{itemize}
\section{Related Work}
\label{sec:related}

\subsection{Sketch Extraction}

At its core, sketch extraction utilizes edge detection. Edge detection serves as the foundation not only for sketch extraction but also for tasks like object detection and segmentation~\cite{zhang2015level,arbelaez2010contour}. Initial edge detection studies primarily focused on identifying edges based on abrupt variations in color or brightness~\cite{canny1986computational,winnemoller2011xdog}. Although these techniques are direct and efficient without requiring extensive datasets to train on, they often produce outputs with artifacts, like scattered dots or lines. 

To make extracted sketches authentic, learning-based strategies have been introduced. These strategies excel at identifying object borders or rendering lines in distinct styles~\cite{a2s,xie2015holistically,lllyasviel2017sketchKeras,li2019photo,li-2017-deep}. \citet{chan2022learning} took a step forward from prior techniques by incorporating the depth and semantic information of images to procure superior-quality sketches. In a more recent development, Ref2sketch~\cite{ashtari2022reference} permits to extract stylized sketches using reference sketches through paired training. Semi-Ref2sketch~\cite{seo2023semi} adopted contrastive learning for semi-supervised training. All of these methods share the same limitation; they require a large amount of sketch data for training, which is hard to gather. Due to data scarcity, training a sketch extraction model is generally challenging. To address this challenge, our method is designed to train a sketch generator using just one manual drawing.

\subsection{Diffusion Features for Downstream Task}

Diffusion models \cite{ho2020denoising,nichol2021improved} have shown cutting-edge results in tasks related to generating images conditioned on text prompt~\cite{rombach2022high,saharia2022photorealistic,ramesh2021zero}. There have been attempts to analyze the features for utilization in downstream tasks such as segmentation~\cite{baranchuk2021label,xu2023open,khani2023slime}, image editing~\cite{tumanyan2023plug}, and finding dense semantic correspondence~\cite{luo2023diffusion,zhang2023tale,tang2023emergent}. Most earlier studies chose a specific subset of features for their own downstream tasks. Recently, \citet{luo2023diffusion} proposed an aggregator that learns features from all layers and that uses equally sampled time steps. We advance a step further by analyzing and selecting the features from multiple timesteps, which represent the overall features. We also propose a two-stage aggregation network and feature-fusing decoder utilizing additional information from VAE to generate finer details.

\subsection{Deep Features for Sketch Extraction}

Most of recent sketch extraction methods utilize the deep features of a pretrained model for sketch extraction training~\cite{ashtari2022reference,seo2023semi,yi2019apdrawinggan,yi2020unpaired}. While the approach of utilizing deep features from a pretrained classifier~\cite{johnson2016perceptual,zhang2018unreasonable} is widely used to measure perceptual similarity, vision-language models such as CLIP~\cite{radford2021learning} are used to measure semantic similarity~\cite{chan2022learning,vinker2022clipasso}. These methods indirectly use the features by comparing them for the loss calculation during the training process instead of using them directly to generate a sketch. Unlike the previous approaches, we directly use the denoising diffusion features that contain rich information to extract sketches for the first time.
\section{Diffusion Features}
\begin{figure*}[htb]
  \centering
  \hspace{-5mm}
  \includegraphics[width=1.05\linewidth]{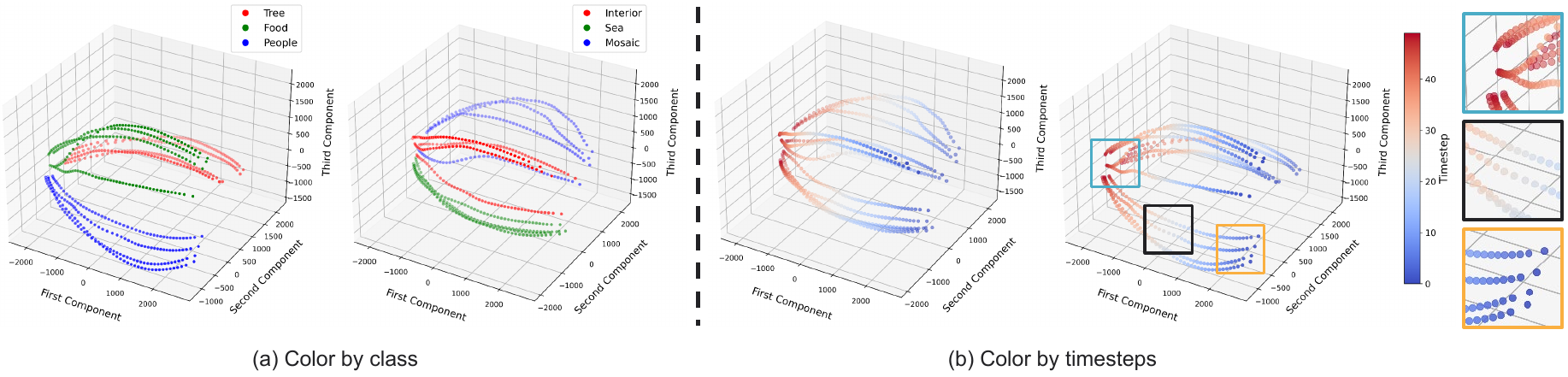}
  \caption{Analysis on sampled features. PCA is applied to DDIM sampled features from different classes. (a) : features colored with human-labeled classes. (b) : features colored with denoising timesteps.}
  \vspace{-3mm}
   \label{fig:pca}
\end{figure*}

During a backward diffusion process, a latent or image with noise repeatedly invokes a UNet~\cite{ronneberger2015u} to reduce the noise. The UNet produces several intermediate features with different shapes. This collection of features contains rich information about texture and semantics, which can be used to generate an image in various domains. For instance, features from the lower to intermediate layers of the UNet reveal global structures and semantic regions, while features from higher layers exhibit fine and high-frequency information \cite{tumanyan2023plug,luo2023diffusion}. Furthermore, features become more fine-grained over time steps~\cite{hertz2022prompt}. As these features have different information depending on their embedded layers and processed timesteps, it is important to select diverse features to fully utilize the information they provide.

\subsection{Diffusion Features Selection}\label{sec:selection}
Here, we first present a method for selecting features by analysis. Our approach involves selecting representative features from all the denoising timesteps and building our novel sketch generator, $G_{sketch}$ to extract a sketch from an image by learning from a single data. To perform analysis for this purpose, we first sampled 1,000 images randomly and collected all the features from multiple layers and timesteps during Denoising Diffusion Implicit Model (DDIM) sampling, with a total of 50 steps~\cite{song2020denoising}.

We conducted Principal component analysis (PCA) on these features from multiple classes and all timesteps to examine the distribution of features depending on their semantics and timesteps. The PCA results are visualized in Figure~\ref{fig:pca}. For our experiments, we manually classified the sampled images and their corresponding features into 17 classes with human perception, where each class contains more than 5 images. As illustrated by the left graphs in Figure~\ref{fig:pca} (a), features from the same class tend to have similar characteristics, which can be seen as an additional proof to the previous literature finding that features contain semantic information~\cite{zhang2023tale,baranchuk2021label,xu2023open}. There is also a smooth trajectory across timesteps as shown in Figure~\ref{fig:pca} (b). Therefore, selecting features from a hand-crafted interval can be more beneficial than using a single feature, as it provides richer information, as previously suggested~\cite{luo2023diffusion}. Upon further examination, we can observe that features tend to start at a similar point in their initial timesteps ($t\approx 50$) and diverge thereafter (cyan box). In addition, during the initial steps, nearby values do not show a large difference compared to those in the middle (black box), while the final features exhibit distinct values even though they are on the same trajectory (orange box).

These findings provide insights that can guide the selection of representative features. As we aim to capture the most informative features across timesteps instead of using all features, we first conducted a K-means cluster analysis (K-means)~\cite{hotelling1933analysis} with Within Clusters Sum of Squares distance (WCSS) to determine the number of representative clusters. One way to compute the K-means cluster with WCSS distance is to use the elbow method. However, we could not identify a clear visual elbow when 30 PCA components were used. Therefore, we used a combination of the Silhouette Score (SS) \cite{rousseeuw1987silhouettes} and the Davies-Bouldin Index (DBI) \cite{davies1979cluster}. For all features from each sampled image, we chose the first $K$ that matched both $k'$'th highest SS score and $k'$'th lowest DBI score. 

From this process, we chose our $K$ as 13 although this $K$ value may vary with the number of diffusion sampling processes. We select the representative features from the center of each cluster to use them as input to our sketch generation network. To verify that the selected features indeed offer better representation compared to those selected from equal timesteps and random features, we calculated the minimum Euclidean distance from each projected feature to the selected 13 features across 1,000 images. We found that our method led to the minimum distance (18,615.6) among the distances achieved by using the equal timestep selection (19,004.9) and random selection (23,957.2). More explanations are provided in the supplementary material.

\subsection{Diffusion Features Aggregation}
\label{sec:aggregation}

Inspired by feature aggregation networks for downstream tasks \cite{xu2023open,luo2023diffusion}, we build our two-level aggregation network and feature fusing decoder (FFD), both of which constitute our new sketch generator ${G}_{sketch}$. The architectures of $G_{sketch}$ and FFD are shown in Figure~\ref{fig:overview} (b) and (d), respectively. The diffusion features $f_{l,t}$, generated on layer $l$ and timestep $t$, are passed through the representative feature gate $G^*$. They are then upsampled to a certain resolution by $U_{md}$ and $U_{tp}$, and passed through a bottleneck layer $B_l$ followed by being assigned with mixing weights $w$. The second aggregation network receives the first fused feature $F_{fst}$ as an additional input feature. 
\begin{equation}
{\footnotesize
\begin{aligned}
F_{fst} &= \sum_{t=0}^{T} \sum_{l=1}^{l_{md}} w_{l,t} \cdot B_l(U_{md}(G^*(f_{l,t}))), \\
F_{fin} &= \sum_{t=0}^{T} \sum_{l={l_{md}+1}}^{L} w_{l,t} \cdot B_l(U_{tp}(G^*(f_{l,t}))) \\
&+\sum_{l={l_{md}+1}}^{L} w_{l} \cdot B_l(U_{tp}(F_{fst}))
\end{aligned}
}
\end{equation}
Here, $L$ is the total number of UNet layers, while $l_{md}$ indicates the middle layer, which are set to be 12 and 9, respectively. Bottleneck layer $B_l$ is shared across timesteps. $T$ is the total number of timesteps. $F_{fst}$ denotes the first level aggregated features and $F_{fin}$ denotes the final aggregated features. \fixed{These two levels of aggregation allow us to utilize the features in a memory efficient manner by mixing the features sequentially in a lower resolution first and then in a higher resolution.}

\subsection{VAE Decoder Features}
\label{sec:vae}

\begin{figure}[htb]
  \centering
  \includegraphics[width=0.8\linewidth]{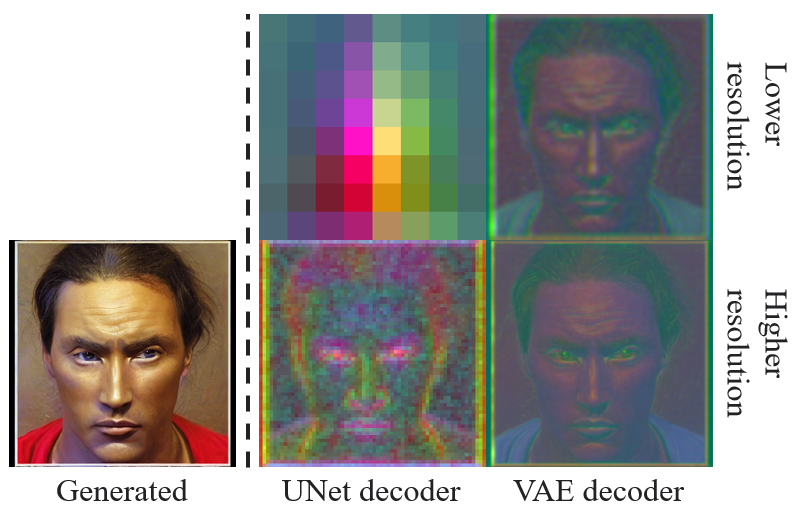}
  \vspace{-3mm}
  \caption{Visualization of features from UNet and VAE in lower and higher resolution layers. Lower resolution layers are the first layers while higher resolution layers are the 11th for UNet and the 9th for VAE.}
  \vspace{-3mm}
   \label{fig:features}
\end{figure}

Unlike recent applications on utilizing diffusion features, where semantic correspondences are more important than high-frequency details, sketch generation utilizes both semantic information and high-frequency details such as texture. As shown in Figure~\ref{fig:features}, VAE decoder features contain high-frequency details such as hair and wrinkles. From this observation, we designed our network to utilize VAE features following the aggregation of UNet features. Extended visualizations are provided in the supplementary material.

\fixed{We utilize all the VAE features from the residual blocks to build FFD. The aggregated features $F_{fin}$ and VAE features are fused together to generate the output sketch. Specifically, in the fusing step $i$, 
VAE features with the same resolution are passed through the channel reduction layer followed by the convolution layer. These processed features are concatenated to the previously fused feature $x_{i}$ and the result is passed through the fusion layer to output $x_{i+1}$. For the first step ($i=0$), $x_0$ is $F_{fin}$. All features in the same step has same resolution. We denote the number of total features at $i$ as $N$ without subscript for simplicity. This process is shown in Figure~\ref{fig:overview} (d) and can be expressed as follows:}

\vspace{-3mm}
\begin{equation}
{\small
\begin{aligned}
x_{i+1} &= \text{FUSE}\left[ \left\{ \sum_{n=1}^{N} \text{Conv}(\text{CH}(v_{i,n}))\right\} + x_{i} \right] \\
\hat{I}_{sketch} &= \text{OUT}\left[ \left\{\sum_{n=1}^{N} \text{Conv}(\text{CH}(v_{M,n}))\right\} + x_{M} + {I}_{source} \right] 
\end{aligned}
}
\end{equation}

\fixed{where $CH$ is the channel reduction layer, Conv is the convolution layers, FUSE is the fusion layer, OUT is the final convolution layer applied before outputting a $\hat{I}_{sketch}$, $\sum$ and addition represent concatenation in the channel dimension. Only at the last step ($i=M$), the source image, $I_{source}$ is also concatenated to generate the output sketch.}
\section{DiffSketch}
\begin{figure*}[htb]
  \centering
  \includegraphics[width=\linewidth]{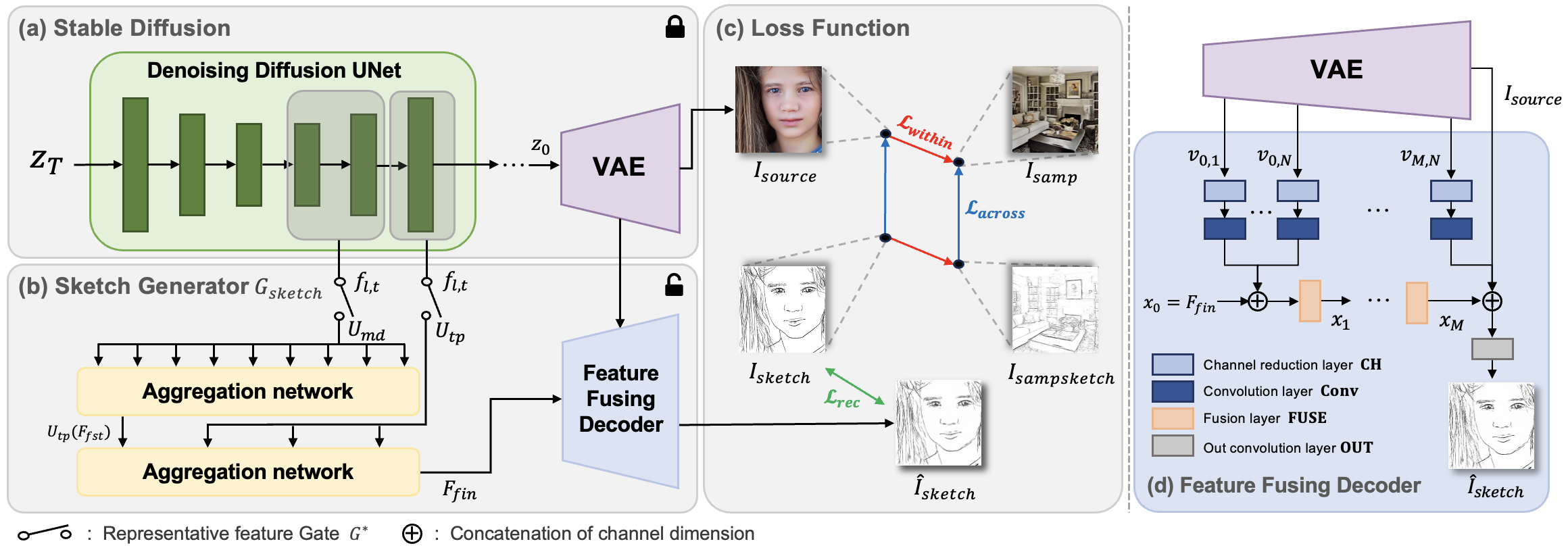}\vspace{-1.0em}
  \caption{Overview of Diffsketch. The UNet features generated during the denoising process are fed to the Aggregation networks to be fused with the VAE features to generate a sketch corresponding to the image that Stable Diffusion generates.}
  \vspace{-1.0em}
   \label{fig:overview}
\end{figure*}

DiffSketch learns to generate a pair of image and sketch through the process described below, which is also shown in Figure~\ref{fig:overview}.

\begin{enumerate}  

\item First, the user generates an image using a prompt with Stable Diffusion (SD)~\cite{rombach2022high} and draws a corresponding sketch while its diffusion features $F$ are kept.

\item The diffusion features $F$, its corresponding image $I_{source}$, and drawn sketch $I_{sketch}$ constitute a triplet data to train the sketch generator $G_{sketch}$ with directional CLIP guidance. 

\item With trained $G_{sketch}$, paired image and sketch can be generated with a condition. This becomes the input for the distilled network for fast sketch extraction.

\end{enumerate}
In the following subsections, we will describe the structure of sketch generator $G_{sketch}$ (Sec.~\ref{subsec:network}), its loss functions (Sec.~\ref{subsec:training}), and the distilled network (Sec.~\ref{subsec:distil}).  

\subsection{Sketch Generator}
\label{subsec:network}

Our sketch generator $G_{sketch}$ is built to utilize the features from the denoising diffusion process by performed UNet and the VAE as described in Secs.~\ref{sec:aggregation} and ~\ref{sec:vae}. \fixed{
$G_{sketch}$ takes the representative features from UNet as input, and aggregate them and fuse them} with the VAE decoder features ${v}_{i,n}$ to synthesizes the corresponding sketch $\hat{I}_{sketch}$. Unlike other image-to-image translation-based sketch extraction methods in which the network takes an image as input~\cite{ashtari2022reference,chan2022learning,seo2023semi}, our method accepts multiple deep features that have different spatial resolutions and channels.

\subsection{Objectives}
\label{subsec:training}

To train $G_{sketch}$, we utilize the following loss functions:

\begin{equation}
\label{eq:total}
L = L_{\text{rec}} +  \lambda_{\text{across}} L_{\text{across}} + \lambda_{\text{within}} L_{\text{within}}
\end{equation}

where $\lambda_{\text{across}}$ and $\lambda_{\text{within}}$ are the balancing weights. $L_{\text{across}}$ and $L_{\text{within}}$ are directional CLIP losses proposed in Mind-the-gap (MTG)~\cite{zhu2022mind}, where $L_{\text{within}}$ preserves the direction across the domain, by enforcing the difference between $I_{samp}$ and $I_{source}$ to be similar to that between $I_{sampsketch}$ and $I_{sketch}$ in CLIP embedding space. Similarly, $L_{\text{across}}$ enforces the difference between $I_{sampsketch}$ and $I_{samp}$ to be similar to that between $I_{sketch}$ and $I_{source}$. $L_{\text{rec}}$ enforces the generated sketch from one known feature $F$ and the ground truth sketch $I_{sketch}$ to be similar. While MTG uses an MSE loss for the pixel-wise reconstruction, we use an L1 distance to avoid blurry sketch results, which is important in the generation of stylized sketches. Our $L_{\text{rec}}$ can be expressed as follows:
\begin{equation}
\label{eq:recon}
L_{\text{rec}} = \lambda_{\text{L1}} L_{\text{L1}} +  \lambda_{\text{LPIPS}} L_{\text{LPIPS}} + \lambda_{\text{CLIPsim}} L_{\text{CLIPsim}}
\end{equation}

where $\lambda_{\text{L1}}$,  $\lambda_{\text{LPIPS}}$, and $\lambda_{\text{CLIPsim}}$ are the balancing weights. $L_{\text{CLIPsim}}$ calculates the semantic similarity in the cosine distance,  $L_{\text{LPIPS}}$~\cite{zhang2018unreasonable} captures the perceptual similarity, and $L_{\text{L1}}$ calculates the pixel-wise reconstruction. More details can be found in Sec.~\ref{sec:Implementation}.

\subsection{Sampling Scheme for Training}
\label{subsec:condition}
\fixed{Our method uses one source image and its corresponding sketch as the only ground truth when guiding the sketch style, using the direction of CLIP embeddings.} Therefore, our losses rely on well-constructed CLIP manifold. When the domains of two images $I_{source}$ and $I_{samp}$ differ largely, the confidence in the directional CLIP loss becomes low in general (experiment details are provided in the supplementary material). To fully utilize the capacity of the diffusion model and produce sketches in diverse domains, however, it is important to train the model on diverse examples.

To ensure learning from diverse examples without decreasing the CLIP loss confidence, we propose a novel sampling scheme, condition diffusion sampling for training (CDST). We envision that this sampling can be useful when training a model with a conditional generator. This method initially samples a data $I_{samp}$ from one known condition $C$ and gradually changes the sampling distribution to random by using a diffusion algorithm when training the network. The condition on the iteration $iter$ ($ 0 \leq iter \leq S $) can be described as follows:

\begin{equation}
\begin{aligned}
\alpha_{iter} &= \sqrt{(1 - \frac{iter}{S})},  \beta_{iter} = \sqrt{\frac{iter}{S}},      \\
C_{iter} &= \frac{\alpha_{iter}}{\alpha_{iter} +\beta_{iter}} C + \frac{\beta_{iter}}{\alpha_{iter} + \beta_{iter}} D_{SD},
\end{aligned}
\end{equation}\label{CDST}
where $D_{SD}$ represents the distribution of the pretrained SD, while $S$ indicates the number of total diffusion duration during training.

\subsection{Distillation}
\label{subsec:distil}

Once the sketch generator $G_{sketch}$ is trained, DiffSketch can generate pairs of images and sketches in the trained style. This generation can be performed either randomly or with a specific condition. Due to the nature of the denoising diffusion model, however, in which the result is refined through the denoising process, long processing time and high memory usage are required. Moreover, when extracting sketches from images, the quality can be degraded because of the inversion process. Therefore, to perform image-to-sketch extraction efficiently while ensuring high-quality results, we train $\text{DiffSketch}_{distilled}$ using Pix2PixHD~\cite{wang2018pix2pixHD}.

To train $\text{DiffSketch}_{distilled}$, we extract 30k pairs of image and sketch samples using our trained DiffSketch, adhering to CDST. Additionally, we employ regularization to ensure that the ground truth sketch $I_{sketch}$ can be generated and discriminated effectively during the training of $\text{DiffSketch}_{distilled}$. With this trained model, images can be extracted in a given style much more quickly than with the original DiffSketch.

\begin{table*}[t]
\centering
\renewcommand{\arraystretch}{0.95} 
\setlength{\tabcolsep}{6pt}
\caption{Quantitative results on ablation with LPIPS and SSIM. Best scores are denoted in bold.} \vspace{-10pt}
\small
\begin{tabular}{|c|cc|cc|cc|cc|}
\noalign{\smallskip}\noalign{\smallskip}\hline
Sketch Styles & \multicolumn{2}{c|}{anime-informative} & \multicolumn{2}{c|}{HED} & \multicolumn{2}{c|}{XDoG} & \multicolumn{2}{c|}{\textbf{Average}} \\ \hline
Methods & LPIPS↓ & SSIM↑ & LPIPS↓ & SSIM↑ & LPIPS↓ & SSIM↑ & LPIPS↓ & SSIM↑\\
\hline
Ours & 0.2054 & 0.6835 & \textbf{0.2117} & \textbf{0.5420} & \textbf{0.1137} & 0.6924 & \textbf{0.1769} & \textbf{0.6393} \\
Non-representative features 1 & 0.2154 & 0.6718 & 0.2383 & 0.5137 & 0.1221 & 0.6777 & 0.1919 & 0.6211 \\
Non-representative features 2 & 0.2042 & 0.6869 & 0.2260 & 0.5281 & 0.1194 & 0.6783 & 0.1832 & 0.6311 \\
One timestep features (t=0) & 0.2135 & 0.6791 & 0.2251 & 0.5347 & 0.1146 & \textbf{0.6962} & 0.1844 & 0.6367 \\
W/O CDST & \textbf{0.2000} & \textbf{0.6880} & 0.2156 & 0.5341 & 0.1250 & 0.6691 & 0.1802 & 0.6304 \\
W/O L1 & 0.2993 & 0.3982 & 0.2223 & 0.5011 & 0.1203 & 0.6547 & 0.2140 & 0.5180 \\
FFD W/O VAE features & 0.2650 & 0.5044 & 0.2650 & 0.4061 & 0.2510 & 0.3795 & 0.2603 & 0.4300 \\

\hline
\end{tabular}
\label{Tab:data_ablation_last100}
\end{table*}

\section{Experiments}

\subsection{Implementation Details}
\label{sec:Implementation}
We implemented DiffSketch and trained generator $G_{sketch}$ on an Nvidia V-100 GPU for 1,200 iterations. When training DiffSketch, we applied CDST with $S$ in Eq.~\ref{CDST} to be 1,000. The model was trained with a fixed learning rate of 1e-4. The balancing weights  $\lambda_{\text{across}}$, $\lambda_{\text{within}}$,  $\lambda_{\text{L1}}$, $\lambda_{\text{LPIPS}}$, and $\lambda_{\text{CLIPsim}}$ are fixed at 1, 1, 30, 15, and 30, respectively. $\text{DiffSketch}_{distilled}$ was trained on two A6000 GPUs using the same architecture and parameters from its original paper except for the output channel, where ours was set to one. We also added regularization on every 16 iterations. $\text{DiffSketch}_{distilled}$ was trained with 30,000 pairs that were sampled from DiffSketch with CDST ($S=30,000$).

LPIPS~\cite{zhang2018unreasonable} and SSIM~\cite{wang2004image} were used for evaluation metrics, in both ablation study and comparison with baselines. LPIPS was to calculate perceptual similarity with pre-trained classifier. SSIM was calculated for structural similarity of sketch image.

\subsection{Datasets}
For training, DiffSketch requires a sketch corresponding to an image generated from SD. To facilitate a numerical comparison, we established the ground truth for given images. Specifically, three distinct styles were employed for quantitative evaluation:
1) HED~\cite{xie2015holisticallynested} utilizes nested edge detection and is one of the most widely used edge detection methods.
2) XDoG~\cite{winnemoller2012xdog} takes an algorithmic approach of using a difference of Gaussians to extract sketches.
3) Informative-anime~\cite{chan2022learning} employs informative learning. This method is the state-of-the-art among single modal sketch extraction methods and is trained on the Anime Colorization dataset~\cite{kim2018animesketch}, which consists of 14,224 sketches. For qualitative evaluation, we added hand-drawn sketches of two more styles. 

For testing, we employed the test set from BSDS500 dataset~\cite{martin2001database} and also randomly sampled an additional 1,000 images from the test set of Common Objects in Context (COCO) dataset~\cite{lin2015microsoft}. As a result, our training set consisted of 3 sketches and the test dataset consisted of 3,600 pairs (1,200 pairs for each style) of image-sketch. Two hand-drawn sketches were used only for perceptual study because there is no ground truth to compare with.

\subsection{Ablation Study} \label{subsec:ablation}
\begin{figure*}[htb]
  \centering
  \includegraphics[width=0.9\linewidth]{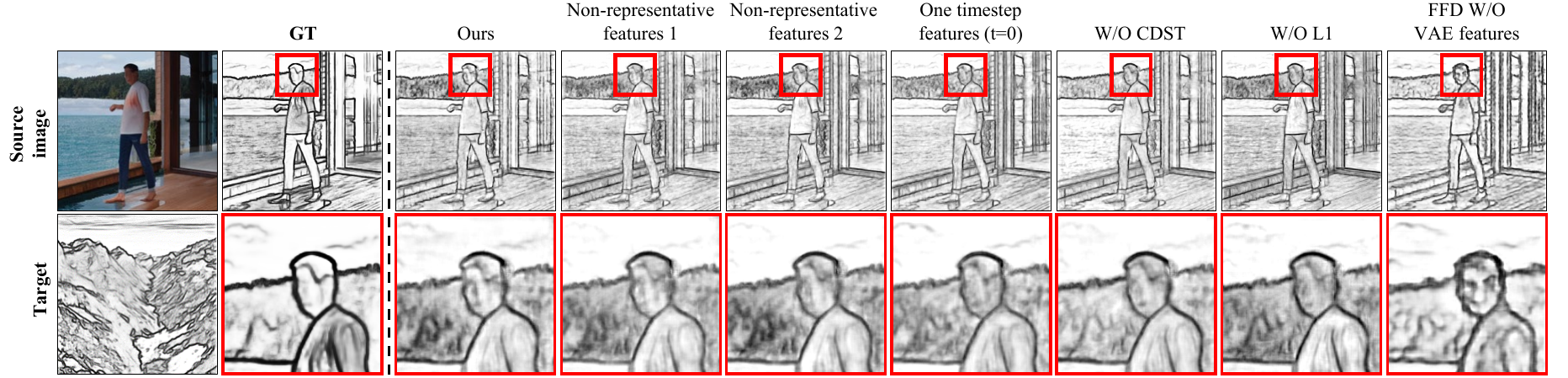}\vspace{0em}
  \caption{Visual examples of the ablation study. Ours generates higher quality results with details such as face, separated with hair region, compared to the alternatives.}
  \vspace{-1.0em}
   \label{fig:ablation}
\end{figure*}
\begin{figure}[htb]
  \centering
  \includegraphics[width=\linewidth]{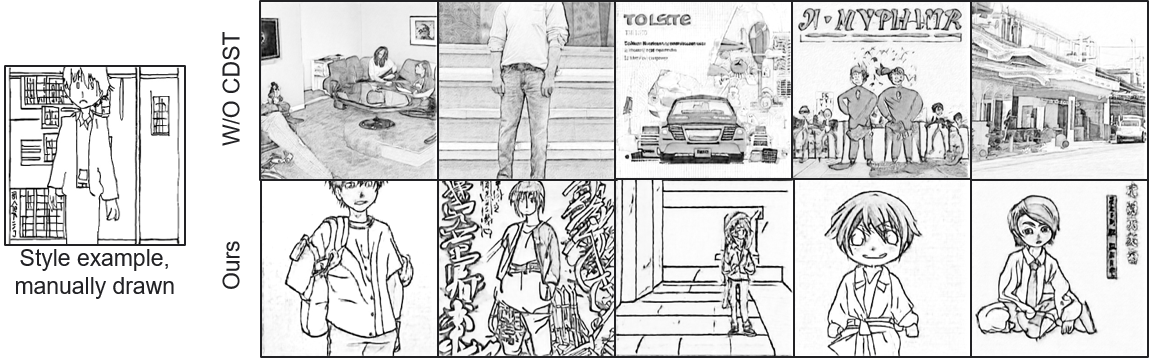}\vspace{0em}
  \caption{Visualization of additional ablation: Ours were trained and sampled with CDST. In contrast, W/O CDST were trained and sampled randomly.}
  \vspace{-1.0em}
   \label{fig:ablation2}
\end{figure}

We conducted an ablation study on each component of our method compared to the baselines as shown in Table~\ref{Tab:data_ablation_last100}. Experiment were performed to verify the contribution of each component; feature selections, CDST, losses, and FFD. To perform the ablation study, we randomly sampled 100 images and extracted sketches with HED, XDog, and Anime-informative and paired them with all 100 images. All seeds were fixed to generate sketches from the same sample.

The ablation study was conducted as follows. For Non-representative features, we randomly selected the features from the denoising timesteps while keeping the number of timesteps equal to ours (13). \fixed{We performed this random selection and analysis twice. For one timestep feature, we only used the features from the final timestep $t=0$. To produce a result without CDST, we executed random text prompt guidance for the diffusion sampling process during training. For the alternative loss approach, we contrasted L1 Loss with L2 Loss for pixel-level reconstruction, as proposed in MTG. To evaluate the effect of the FFD, we produced sketches after removing the VAE features.}

\fixed{The quantitative and qualitative results of the ablation study are shown in Table~\ref{Tab:data_ablation_last100} and Figure~\ref{fig:ablation}, respectively. Ours achieved the higest average scores on both indices. Both Non-representative features achieved overall low scores indicating that representative feature selection helps obtain rich information. Similarly, using one time step features achieved lower scores than ours on average, showing the importance of including diverse features. W/O CDST scored lower than ours on both HED and XDoG styles. W/O L1 and FFD W/O features performed the worst due to the blurry and blocky output, respectively. The blocky results are due to lack of fine information from VAE.}

\paragraph{Condition Diffusion Sampling for Training}

\fixed{While we tested on randomly generated images for quantitative evaluation, our CDST can be applied to both training DiffSkech and sampling for training $\text{DiffSketch}_{distilled}$. Therefore, we performed an additional ablation study on CDST, comparing Ours (trained and sampled with CDST), with W/O CDST (trained and sampled randomly). The outline of the sketch is clearly reproduced, following the style, when CDST is used as shown in Figure~\ref{fig:ablation2}.}

\subsection{Comparison with Baselines}
We compared our method with 5 different alternatives including state-of-the-art sketch extraction methods~\cite{ashtari2022reference,seo2023semi} and diffusion based methods~\cite{ruiz2023dreambooth,kwon2023diffusionbased,gal2022image}. Ref2sketch~\cite{ashtari2022reference} and Semi-Ref2sketch~\cite{seo2023semi} are methods specifically designed to extract sketches in the style of a reference from a large pretrained network on diverse sketches in a supervised (Ref2sketch) and a semi-supervised (Semi-Ref2sketch) manner. DiffuseIT~\cite{kwon2023diffusionbased} is designed for image-to-image translation by disentangling style and content. DreamBooth~\cite{ruiz2023dreambooth} finetunes a Stable Diffusion model to generate personalized images, while Textural Inversion~\cite{gal2023textualinversion} optimizes an additional text embedding to generate a personalized concept for a style or object. For DreamBooth and Textual Inversion, DDIM inversion was conducted to extract sketches.

\begin{figure*}[htb]
  \centering
  \includegraphics[width=1\linewidth]{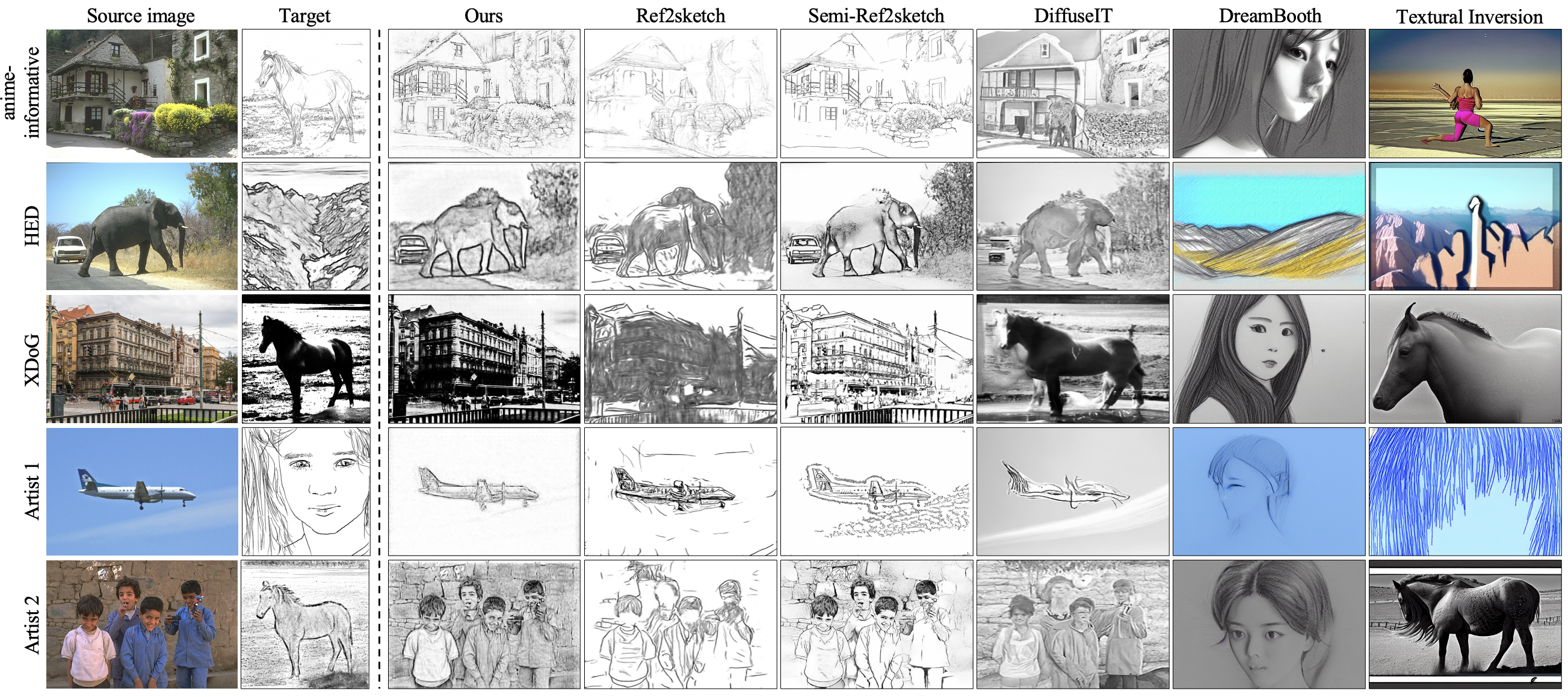}\vspace{-1.0em}
  \caption{Qualitative comparison with alternative sketch extraction methods.}
   \label{fig:comparsion}
\end{figure*}

Table~\ref{Tab:updated_data_experiment} presents the result of the quantitative evaluation that used BSDS500 and COCO datasets in a one-shot setting. Overall, ours achieved the best scores. \fixed{While Semi-Ref2sketch scored higher on some of SSIM scores, the method relies on a large sketch dataset to train while ours requires only one.} Figure~\ref{fig:comparsion} presents visual results produced by different methods. While Semi-Ref2sketch and Ref2sketch generated superior quality sketches to the results produced by others, they do not faithfully follow the style of the reference sketches, especially for dense styles. Diffusion-based methods sometimes overfit to the style image (DiffuseIT) or change the content of the images (DreamBooth, Textual Inversion). $\text{DiffSketch}_{distilled}$ generated superior results compared to these baselines, effectively maintaining its styles and content.

\begin{table*}[t]
\centering
\renewcommand{\arraystretch}{1}
\setlength{\tabcolsep}{8pt}
\vspace{-5pt}
\caption{Quantitative comparison of different methods on BSDS500 and COCO datasets.} \vspace{-5pt}
\footnotesize
\begin{tabular}{|c|cc|cc|cc|cc|}
\hline
& \multicolumn{2}{c|}{BSDS500 - anime} & \multicolumn{2}{c|}{BSDS500 - HED} & \multicolumn{2}{c|}{BSDS500 - XDoG}& \multicolumn{2}{c|}{BSDS500 - average} \\
Methods & LPIPS↓ & SSIM↑ & LPIPS↓ & SSIM↑ & LPIPS↓ & SSIM↑ & LPIPS↓ & SSIM↑\\
\hline
$\text{Ours}_{distilled}$ & \textbf{0.21746} & 0.49343 & \textbf{0.22706} & \textbf{0.59314} & \textbf{0.14280} & \textbf{0.64874} & \textbf{0.19577} & \textbf{0.57844} \\
Ref2sketch & 0.33621 & 0.46932 & 0.41993 & 0.31448 & 0.57096 & 0.13095 & 0.44237 & 0.30492 \\
Semi-Ref2sketch & 0.23916 & \textbf{0.50972} & 0.39675 & 0.34200 & 0.50447 & 0.30918 & 0.38013 & 0.38697 \\
DiffuseIT & 0.48365 & 0.29789 & 0.49217 & 0.19104 & 0.57335 & 0.11030 & 0.51639 & 0.19974 \\
DreamBooth & 0.80608 & 0.30149 & 0.74550 & 0.18523 & 0.72326 & 0.19465 & 0.75828 & 0.22712 \\
Textual Inversion & 0.82789 & 0.26373 & 0.77098 & 0.16416 & 0.64662 & 0.21953 & 0.74850 & 0.21581 \\
\hline
\end{tabular}
\begin{tabular}{|c|cc|cc|cc|cc|}
\hline

& \multicolumn{2}{c|}{COCO - anime} & \multicolumn{2}{c|}{COCO - HED} & \multicolumn{2}{c|}{COCO - XDoG}& \multicolumn{2}{c|}{COCO - average} \\
Methods & LPIPS↓ & SSIM↑ & LPIPS↓ & SSIM↑ & LPIPS↓ & SSIM↑ & LPIPS↓ & SSIM↑\\
\hline
$\text{Ours}_{distilled}$ & \textbf{0.17634} & 0.36021 & \textbf{0.20039} & 0.36093 & \textbf{0.14806} & \textbf{0.38319} & \textbf{0.17493} & 0.36811 \\
Ref2sketch & 0.32142 & 0.50517 & 0.37764 & 0.37230 & 0.56012 & 0.16835 & 0.41973 & 0.34861 \\
Semi-Ref2sketch & 0.21337 & \textbf{0.64732} & 0.32920 & \textbf{0.39487} & 0.47974 & 0.31894 & 0.34077 & \textbf{0.45371} \\
DiffuseIT & 0.46527 & 0.36092 & 0.47905 & 0.24611 & 0.56360 & 0.14595 & 0.50264 & 0.25099 \\
DreamBooth & 0.76399 & 0.30517 & 0.72278 & 0.22066 & 0.67909 & 0.21655 & 0.72195 & 0.24746 \\
Textual Inversion & 0.81458 & 0.29168 & 0.78835 & 0.19952 & 0.63215 & 0.22074 & 0.74503 & 0.23731 \\
\hline
\end{tabular}

\label{Tab:updated_data_experiment}
\end{table*}

\subsection{Perceptual Study}

We conducted a user study to evaluate different sketch extraction methods on human perception. We recruited 45 participants to complete a survey that used test images from two datasets, processed in five different styles, to extract sketches. Each participant was presented with a total of 20 sets of source image, target sketch style, and resulting sketch. Participants were asked to choose one that best follows the given style while preserving the content of the source image. The result should not depend on demographics distribution, therfore we did not focus on group of people as previous sketch studies~\cite{ashtari2022reference,seo2023semi,chan2022learning}. As shown in Table~\ref{tab:perceptual}, our method received the highest scores when compared with the alternative methods. Ours outperformed the diffusion-based methods by a large margin and even received a higher preference rating than the specialized sketch extraction method that was trained on a large sketch dataset.

\begin{table}[t]
\centering
\caption{Results from the user perceptual study given style example and the source image. The percentage indicates the selected frequency.}
\vspace{-1em}
\footnotesize
\renewcommand{\arraystretch}{0.95} 
\label{tab:perceptual}
\begin{tabular}{|l|c|}\hline 
\multicolumn{1}{|c|}{Methods} & User Score     \\ \hline
Ours                         & \textbf{ 68.67\%} \\ 
Ref2sketch                          &  6.00\%\\ 
Semi-Ref2sketch                     &  18.56\%\\ 
DiffuseIT                           & 0.22\%\\
DreamBooth                          & 0.00\%\\
Textual Inversion                   &  0.22\%\\ 
\hline 
\end{tabular}
\vspace{-0.5cm}
\end{table}
\section{Limitation and Conclusion}

We proposed DiffSketch, a novel method to train a sketch generator using representative features and extract sketches in diverse styles. For the first time, we conducted the task of extracting sketches from the features of a diffusion model and demonstrated that our method outperforms previous state-of-the-art methods in extracting sketches. The ability to extract sketches in a diverse style, trained with one example, will have various use cases not only for artistic purposes but also for personalizing sketch-to-image retrieval and sketch-based image editing.

\fixed{We built our generator network specialized for generating sketches by fusing aggregated features with the features from a VAE decoder. Consequently, our method works well with diverse sketches including dense sketches and outlines. Because our method not directly employ a loss function to compares stroke styles, however, it fails to generate highly abstract sketches or pointillism. One possible research direction could involve incorporating a new sketch style loss that does not require additional sketch data, such as penalizing based on stroke similarity in close-ups.}

Although we focused on sketch extraction, our analysis of selecting representative features and the proposed training scheme are not limited to the domain of sketches. Extracting representative feature holds potential to improve applications leveraging diffusion features, including semantic segmentation, visual correspondence, and depth estimation. We believe this research direction promises to broaden the impact and utility of diffusion feature-based applications.

\clearpage
\setcounter{page}{1}
\newcommand{\myparagraph}[1]{\vspace{2pt}\noindent{\bf #1}}
\maketitlesupplementary
\begin{figure*}[h]
  \centering
  \includegraphics[width=0.9\linewidth]{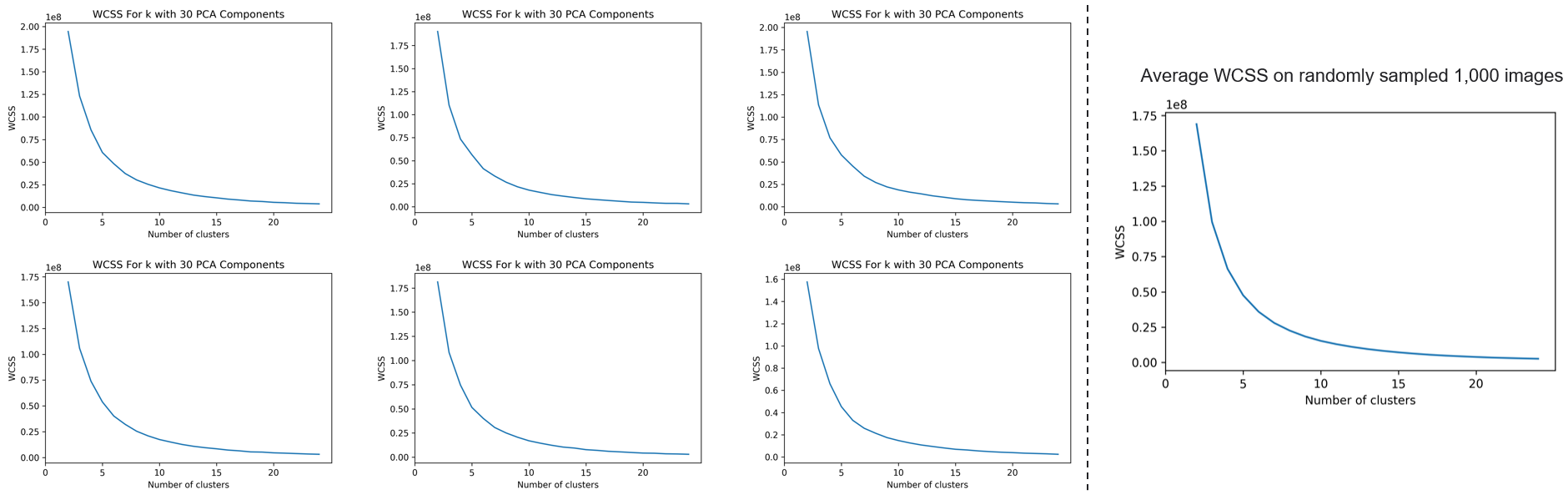}
  \vspace{-3mm}
  \caption{Visualization of WCSS values according to the number used for K-means clustering. The left plots are the WCSS of the features from an randomly sampled image while the right plot shows the average WCSS values of the features from 1,000 randomly sampled images. }

   \label{fig:Kmeans}
\end{figure*}

\section*{Overview}
\label{sec:Overview}
This supplementary material consists of 5 Sections. Section A describes implementation details (Sec.~\hyperref[sec:A]{A}). Sec.~\hyperref[sec:B]{B} provides additional details and findings on diffusion features selection. Sec.~\hyperref[sec:C]{C} presents extended details of VAE decoder features. Sec.~\hyperref[sec:D]{D} contains the results of additional experiments on CDST. Lastly, Sec.~\hyperref[sec:E]{E} presents additional qualitative results with various style sketches.

\section*{A. Implementation Details}
\label{sec:A}

\paragraph{DiffSketch} 
DiffSketch leverages the Stable Diffusion v1.4 sampled with DDIM~\cite{song2020denoising} pretrained with the LAION-5B~\cite{schuhmann2022laion} dataset, which produced images of resolution 512 $\times$ 512. With the pretrained Stable Diffusion, we use a total of 50 time steps T for sampling. The training of DiffSketch was performed for 1200 iterations which required less than 3 hours on an Nvidia V100 GPU. For the training using HED~\cite{xie2015holisticallynested}, we concatenated the first two layers with the first three layers to stylize sketch. In the case of XDoG~\cite{winnemoller2011xdog}, we used Gary Grossi style.

\paragraph{$\text{DiffSketch}_{distilled}$}
$\text{DiffSketch}_{distilled}$ was developed to conduct sketch extraction efficiently with the streamlined generator. The training of $\text{DiffSketch}_{distilled}$ was performed for 10 epochs for 30,000 sketch-image pairs generated from DiffSKetch, following the CDST. The training of $\text{DiffSketch}_{distilled}$ required approximately 5 hours on two Nvidia A6000 GPUs. The inference time of both DiffSketch and $\text{DiffSketch}_{distilled}$ was 4.74 seconds and 0.0139 seconds, respectively, when tested on an Nvidia A5000 GPU with image with same resolutions.

\paragraph{Comparison with Baselines} For the baselines, the settings used in our study were based on the official code provided by the authors and information obtained from their respective papers. For both Ref2Sketch~\cite{ashtari2022reference} and Semi-ref2sketch~\cite{seo2023semi}, we used the given checkpoint, the official pre-trained model provided by the authors. For DiffuseIT~\cite{kwon2023diffusionbased}, we also used the official code and checkpoint given by the authors in which the diffusion model was trained with the Imagenet~\cite{DenDon09Imagenet} dataset, not FFHQ~\cite{karras2019stylegan} because our comparison is not constrained to the face. For Dreambooth~\cite{ruiz2023dreambooth} and Textual Inversion~\cite{gal2023textualinversion}, we used DDIM inversion~\cite{song2020denoising} to invert the source image to the latent code of Stable Diffusion.

\section*{B. Diffusion Features Selection }
\label{sec:B}

To conduct K-means clustering for diffusion feature selection, we first employed the elbow method, visualizing the results. However, a distinct elbow was not visually apparent, as shown in Figure~\ref{fig:Kmeans}. The left 6 images are WCSS values from randomly selected images out of our 1,000 test images. All 6 plots show similar patterns, making it hard to select a definitive elbow as stated in the main paper. The right image, which exhibits similar results, shows the average of WCSS on all 1,000 images. 

\begin{figure*}[h]
  \centering
  \includegraphics[width=\linewidth]{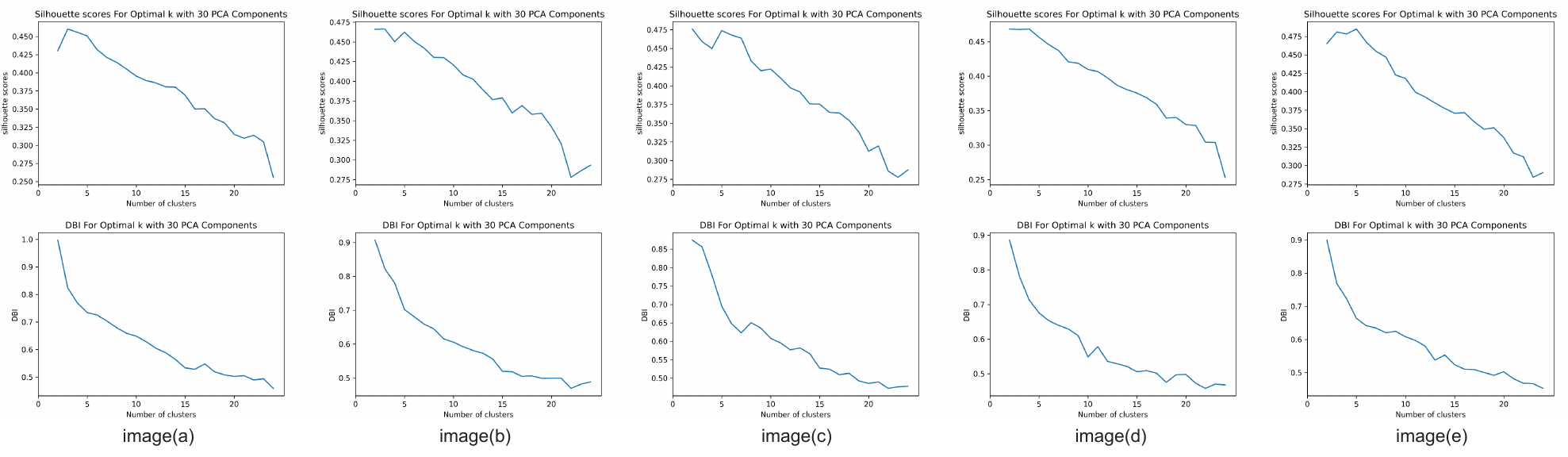}
  \vspace{-3mm}
  \caption{Visualization of contradicting results of Silhouette scores and Davis Bouldin indices on five different images.}

   \label{fig:ssdbi}
\end{figure*}

Therefore, we chose to use the Silhouette score~\cite{rousseeuw1987silhouettes} and Davies-Bouldin index~\cite{davies1979cluster}, which are two of the most widely used numerical method when choosing the optimal number of clusters. However, they are two different methods, whose results do not always match with each other. We first visualized and found the contradicting results of these two methods as shown in Figure~\ref{fig:ssdbi}. Therefore, we chose to use the one that first matches the $i^{\text{th}}$ highest silhouette score and the $i^{\text{th}}$ lowest Davies-Bouldin index simultaneously. This process of choosing the optimal number of clusters can be written as follows :

\begin{algorithm}\label{alg:kmeans}
\caption{Finding the Optimal Number of Clusters}
\begin{algorithmic}[1]
\State $MAX\_clusters = Total\_time\_steps /2$
\State $sil\_indices \gets \text{sorted}(\text{range}(MAX\_clusters), \text{key}=\lambda k: silhouette\_scores[k], \text{reverse}=True)$
\State $db\_indices \gets \text{sorted}(\text{range}(MAX\_clusters), \text{key}=\lambda k: db\_scores[k], \text{reverse}=False)$
\For{$i \gets 0$ \textbf{to} $MAX\_clusters$}
    \If{$sil\_indices[i]$ \textbf{in} $db\_indices[:i+1]$}
        \State $k\_optimal$ = $sil\_indices[i]$+1
        \State \textbf{break}
    \EndIf
\EndFor
\end{algorithmic}
\end{algorithm}

\begin{figure}[h]
  \centering
  \includegraphics[width=\linewidth]{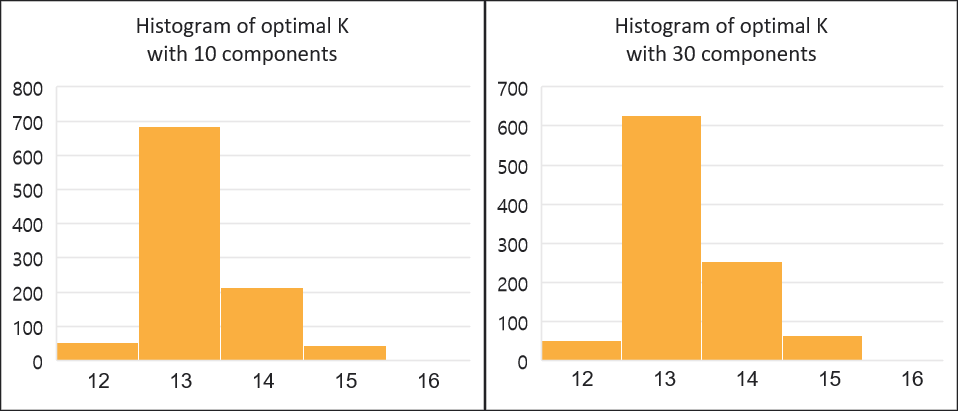}
  \vspace{-3mm}
  \caption{Visualization on histogram for optimal k value with different number of PCA components.}

   \label{fig:hist}
\end{figure}

We conducted this process twice with two different numbers of PCA components (10 and 30), yielding the results shown in Figure~\ref{fig:hist}. The averages (13.26 and 13.34) and standard deviations (0.69 and 0.69) were calculated. As the mode value with both PCA components was 13, and the rounded average was also 13, we chose our optimal k to be 13. Using this number of clusters, we chose the representative feature as the one nearest to the center of each cluster.

From this process, we ended up with the following t values: [0,3,8,12,16,21,25,28,32,35,39,43,47]. To verify the process, if an optimal number of clusters in each image can really be globally adjusted, we compared our selected features against the baselines. These baselines included sampling at equal time intervals (t=[i*4+1 for i in the range of (0,13)]) and randomly selecting 13 values. We calculated the minimum Euclidean distance from each feature and confirmed that our method resulted in the minimum distance across 1,000 randomly sampled images. This is illustrated in Table~\ref{tab:kmeans}.

\begin{table}[h]
\centering
\caption{Sum of the minimum distances from all features}
\small
\setcounter{table}{0}
\renewcommand{\arraystretch}{1.2} 
\label{tab:kmeans}
\begin{tabular}{|l|c|}\hline
\multicolumn{1}{|c|}{Method} & Euclidean Distance \\ \hline
Ours & \textbf{ 18,615.6} \\
Equal time steps & 19,004.9\\
Random sample & 23,957.2\\
\hline
\end{tabular}
\end{table}

\begin{figure*}[htb]
  \centering
  \includegraphics[width=\linewidth]{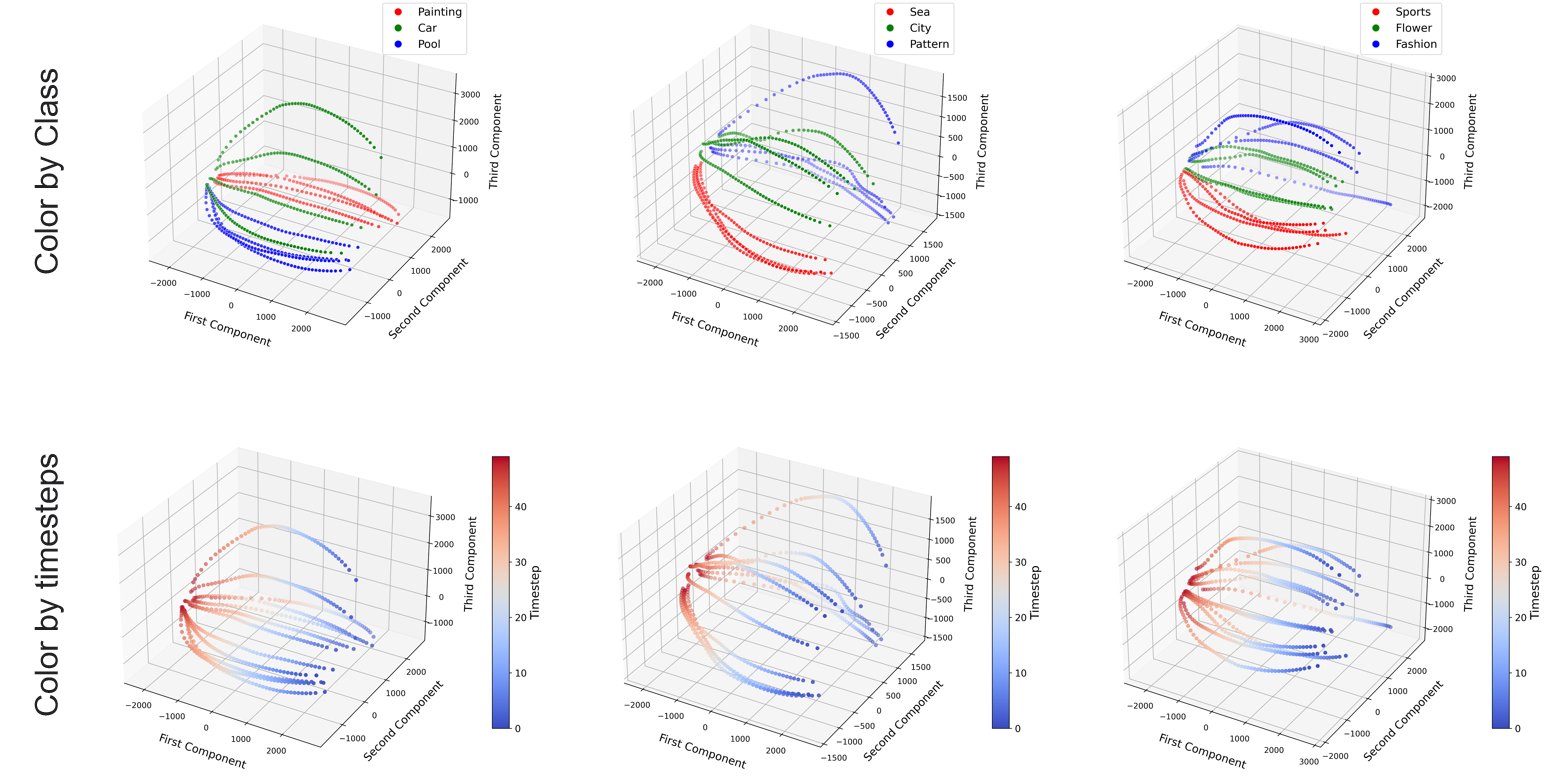}
  \vspace{-3mm}
  \caption{ Additional analysis on sampled features. PCA is applied to DDIM sampled features from different classes. Up : features colored with
human-labeled classes. Down : features colored with denoising timesteps }

   \label{fig:additional_features}
\end{figure*}

In the main paper, we found several key insights through the visualization of features within the manually selected classes, which we summarize extensively here. First, semantically similar images lead to similar trajectories, although not identical. Second, features in the initial stage of the diffusion process (when t is approximately 50) retain similar information despite significant differences in the resulting images. Third, features in the middle stage of the diffusion process (when t is around 25) exhibit larger differences between adjacent features in their time steps. Lastly, the feature at the final time step (t=0) possesses distinctive information, varying significantly from previous values. This is also evident in the additional visualization presented in Figure~\ref{fig:additional_features}.

Our automatically selected features indicate a prioritization of the final feature (t=0), and that selection was made more from the middle stages than from the initial steps (t=[21,25,28] versus t=[43,47]). Our finding offer some guidance for manual feature selecting to consider the time steps, especially when memory is constrained. The order of the preference will on the last feature (t=0), a middle one (t is near 25), and the middle to final time steps while the features from initial steps are preferred less in general. For instance, when selecting four features from 50 time steps, a possible selection could be t=[0,12,25,37].

\subsection*{B.2 Features From Additional Models}
\begin{figure*}[htb]
  \centering
  \includegraphics[width=\linewidth]{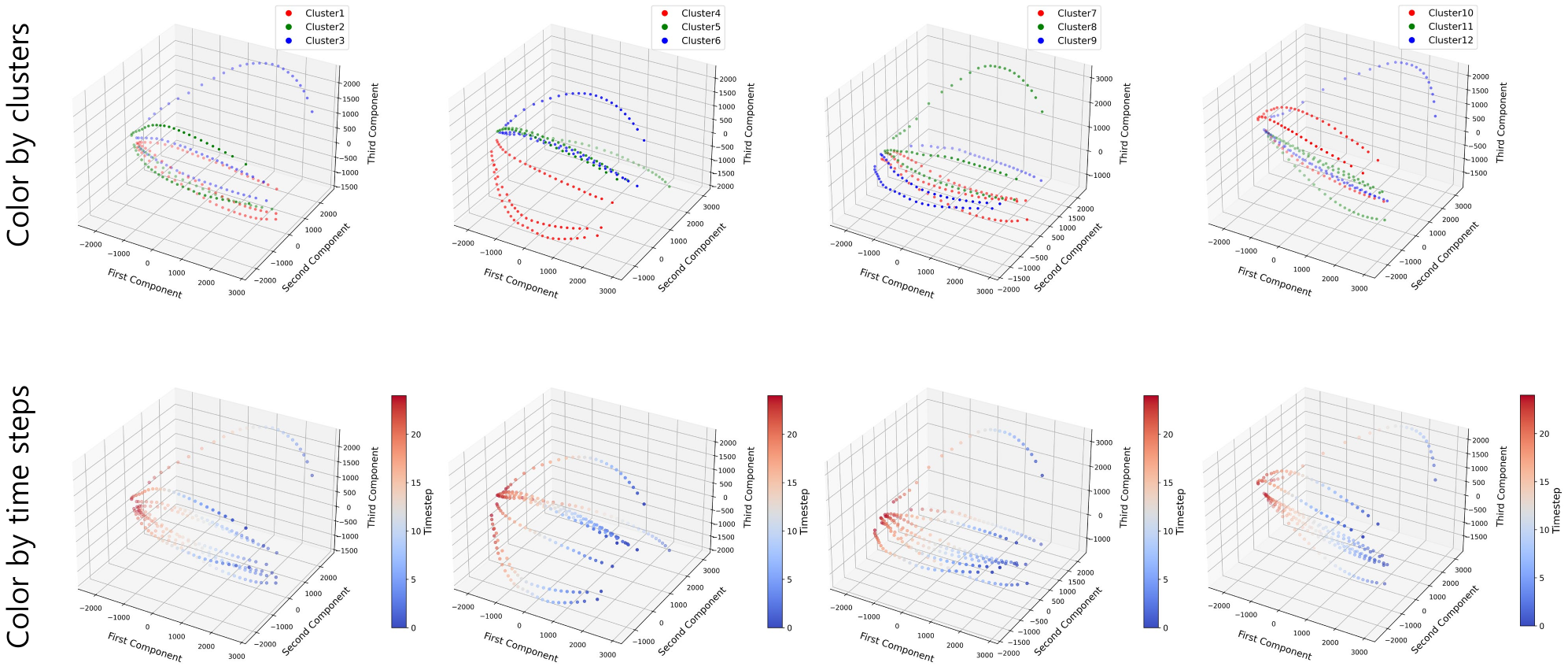}
  \vspace{-3mm}
  \caption{ Additional analysis on sampled features. PCA is applied to 25 steps of DDIM sampled features with different clusters. Up : features colored with DINOv2 clusters. Down : features colored with denoising timesteps. }
   \label{fig:additional_features25}
\end{figure*}

\begin{figure*}[htb]
  \centering
  \includegraphics[width=\linewidth]{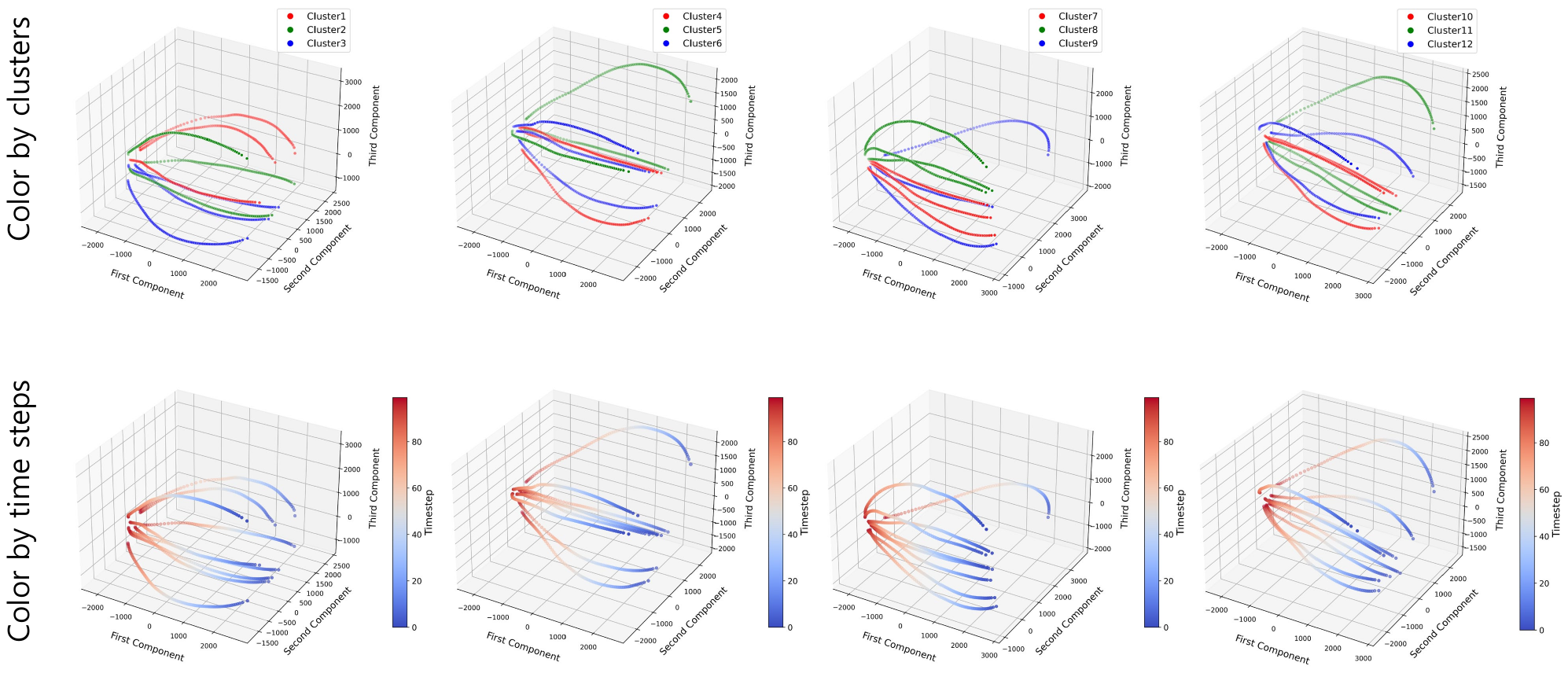}
  \vspace{-3mm}
  \caption{ Additional analysis on sampled features. PCA is applied to 100 steps of DDIM sampled features with different clusters. Up : features colored with DINOv2 clusters. Down : features colored with denoising timesteps.}
   \label{fig:additional_features100}
\end{figure*}

While we focused on T=50 DDIM sampling, for generalization, we examined different intervals (T=25, T=100) and different model. For these experiments, we randomly sampled 100 images. While our main experiments were conducted with manually classified images, we utilized DINOv2~\cite{oquab2023dinov2}, which was contrastively trained in a self-supervised manner and has learned visual semantics. With DINOv2, we separated the data into 15 different clusters and followed the process described in the main paper to plot the features. Here, we used 15 images from each cluster to calculate the PCA axis while we used 17 classes in the main experiments. The results, as shown in Figure~\ref{fig:additional_features25} and Figure~\ref{fig:additional_features100}, indicate that even with different sampling methods, the same conclusions regarding the sampling method can be drawn. The last feature exhibits a distinct value, while the features from the initial time step show similar values.

\begin{figure*}[htb]
  \centering
  \includegraphics[width=\linewidth]{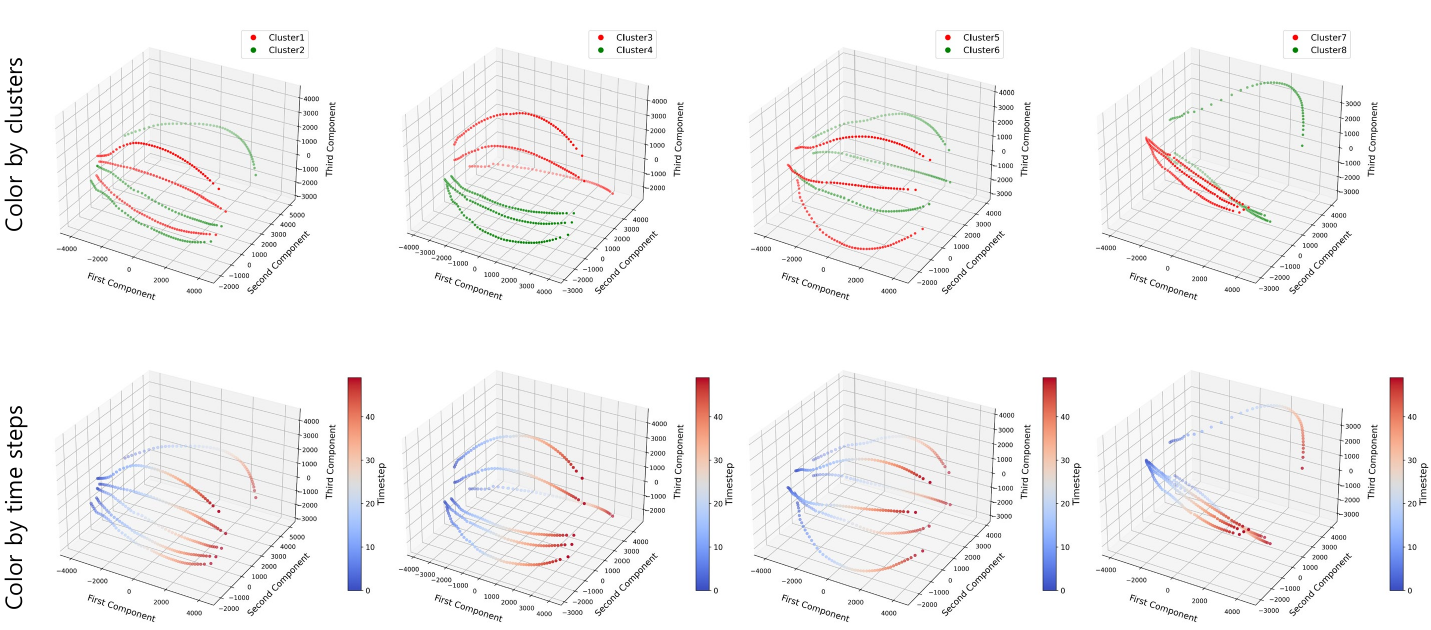}
  \vspace{-3mm}
  \caption{ Additional analysis on Stable diffusion v2.1 sampled features. PCA is applied to 50 steps of DDIM sampled features with different clusters. Up : features colored with DINOv2 clusters. Down : features colored with denoising timesteps.}
   \label{fig:additional_features_sd21}
\end{figure*}

In addition, we also tested on different model, Stable diffusion V2.1 which produce 768$\times$768 images. Following the same process, we randomly sampled 100 images and clustered with DINOv2 and plot as shown in Figure~\ref{fig:additional_features_sd21}. This result also shows that even with different model with different resolution, the same conclusions can be drawn, showing the scalability of our analysis.


\section*{C. VAE Decoder Features}
\label{sec:C}
\begin{figure*}[htb]
  \centering
  \includegraphics[width=\linewidth]{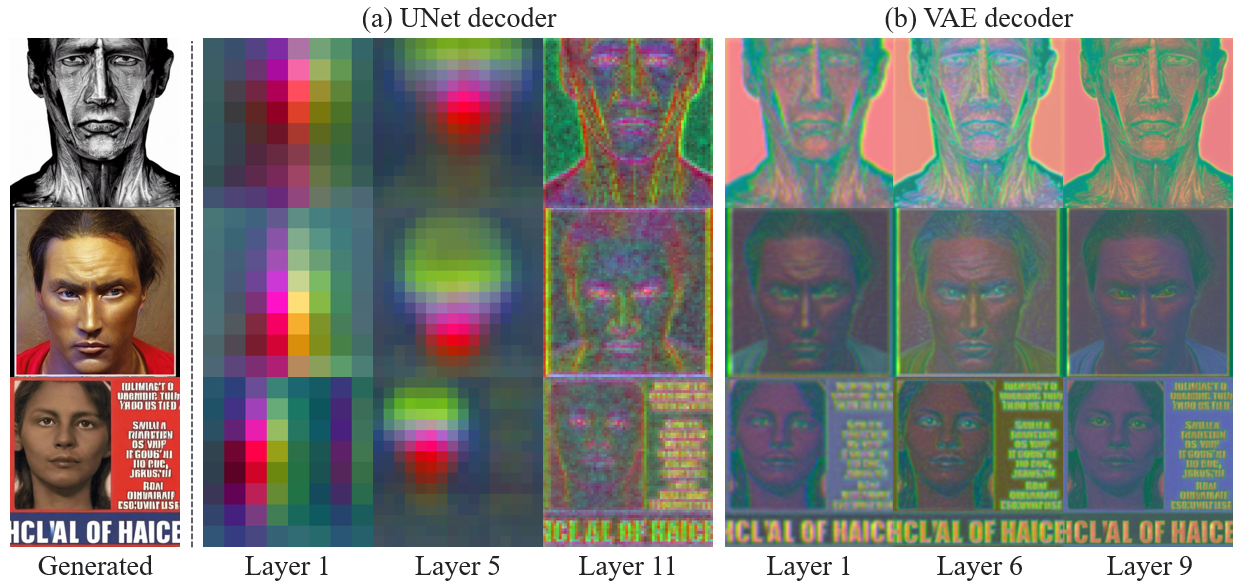}
  \vspace{-3mm}
  \caption{Extended visualization of features from UNet and VAE. (a) shows the UNet decoder features in lower resolution (layers 1), intermediate resolution (layers 5), and higher resolution (layers 11). (b) shows the VAE decoder features in lower resolution (layers 1), intermediate resolution (layers 6), and higher resolution (layers 9).
  }

   \label{fig:sup_vae1}
\end{figure*}

VAE features fused with the Aggregation network features for FFD in the proposed model architecture. Figure~\ref{fig:sup_vae1} shows a visualization of the VAE features. We used a set of 20 generated face images and extracted features from different decoder layers of the UNet and VAE decoder, at the last time step (t=0) similar to that of PNP~\cite{tumanyan2023plug}. We observe that the use of VAE decoder resulted in higher-frequency details than the UNet decoder. While the feature from UNet decoder contains semantic information, the features from VAE decoder produces finer details such as hair, wrinkles, and small letters.

\section*{D. Condition Diffusion Sampling for Training  }
\label{sec:D}

\subsection*{D.1 Rationale Behind CDST}
An underlying assumption of CDST is that for a directional CLIP loss, two images with a similar domain ($I_{source}$ and $I_{samp}$ in the main paper) leads to higher confidence compared to two images with a different domain. To examine this, we performed a \textit{confidence score} test using 4SKST~\cite{seo2023semi} which consists of four different sketch styles paired with color images. 4KST is suitable for the confidence score test because it contains images from two different domains, photos, and anime images, in four different styles.

We manually separated into photos and anime images since it was not labeled. Here, we computed a confidence score to determine if the directional clip loss is more reliable when the calculated source images are in the same domain. We performed a test with three settings, measuring cosine similarity between the images $I_A$ (Photo) and $I_B$ (Anime) from different domains with the corresponding sketches $S_A$ and $S_B$. All these images were encoded into the CLIP embedding. We employed two similarity scores $Sim_{within}$ and $Sim_{across}$ in the same manner as the main paper (Sec.4.2). We calculated the similarity of the features in the photo domain, in the anime domain, and across the two domains. The equation can be expressed as follows:

\begin{table*}[h] 
\setcounter{table}{4}
\centering
\caption{Confidence scores on 4SKST with four different styles.}
\renewcommand{\arraystretch}{1.2} 
\setlength{\tabcolsep}{12pt}
\label{tab:con_sim}
\begin{tabular}{|l|c|c|c|c||c|}\hline 
\multicolumn{1}{|c|}{Similarity} & Style1 & Style2 & Style3 & Style4 & Average \\ \hline
$\textit{confidence(Anime,Anime)}$ & {104.2608} & {102.8716} & {108.2026} & {101.3530} &  {104.1720} \\ 
$\textit{confidence(Photo,Photo)}$ & {101.9346} & {98.8005} & {102.4516} & {100.5453} & {100.9330} \\ 
$\textit{confidence(Photo,Anime)}$& {94.5036} & {94.0189} & {98.1867} & {92.3874} & {94.7742} \\ 
\hline
\end{tabular}
\end{table*}

\begin{equation} 
\label{eq:ab}
\begin{aligned}
Sim(X,Y) = {\cos(\overrightarrow{I_XI_Y} \cdot \overrightarrow{S_XS_Y})+\cos (\overrightarrow{I_XS_X} \cdot \overrightarrow{I_YS_Y})\over{N}} \\
\end{aligned}
\end{equation}
where $cos(a \cdot b)$ is the cosine similarity and N is the total number of $cos$ calculation. X,Y corresponds to the images in each domain.

With these computed similarities, the confidence score in domain A and domain B can be written as follows where $Sim_(ALL,ALL)$ denotes the average similarity of all images:

\begin{equation}
\label{eq:ab}
\begin{aligned}
\textit{confidence(A,B)} =  {{Sim(A,B)}\over{Sim_(ALL,ALL)}} \times 100 \\
\end{aligned}
\end{equation}

In Table~\ref{tab:con_sim}, we show the \textit{confidence} test results on four different style sketches. For all four styles, calculating the directional CLIP loss in the same domain produced higher \textit{confidence} compared to the \textit{confidence} computed across a different domain. Accordingly, we propose a sampling scheme, CDST to train the generator in the same domain at the initial stage of the training, which leads to higher confidence while widening its capacity in the latter iterations of training.

\subsection*{D.2 Additional Experiment on CDST}

In the main paper, we used $D_{SD}$ for CDST. However, the distribution of the condition of a pretrained stable diffusion network is not known. Therefore, we approximate $D_{SD}$ by randomly sampling 1,000 text prompts from the LAION-400M~\cite{schuhmann2021laion400m}, which is a subset of the trained text-image pairs of the SD model. We then tokenized and embedded these prompts for preprocessing, following the process of the pretrained SD model. We conducted PCA on these 1,000 sampled embeddings to extract 512 principal components. We then checked the normality of the sampled embeddings with all 512 principal component axes using the Shapiro-Wilk test~\cite{shapiro1965analysis} with a significance level of $\alpha =5\%$. 

As a result, 214 components rejected the null hypothesis of normality. This indicates that each of its marginals cannot be assumed to be univariate normal. Next, we conducted the Mardia test~\cite{mardia1970measures, mardia1974applications} with the same 1,000 samples, taking into account skewness and kurtosis to check if the distribution is multivariate. The results failed to reject the null hypothesis of normality with a significance level of $\alpha =5\%$. Therefore, we assumed $D_{SD}$ as a multivariate normal distribution for our sampling during training.
 
We examined whether our calculated distribution of stable diffusion ($D_{SD}$) is similar to the ground truth embedding distribution of LAION-400M. For verification, we sampled 100k data from the embedded LAION-400M as a subset of ground truth. We also sampled same amount of embeddings from the multivariate normal distribution (Ours), univariate normal distribution for each axis, and a uniform distribution between the max and min values of the sampled embedded LAION-400M as a baseline. We used Earth moving distance (EMD) \cite{levina2001earth} and found out that the multivariate normal distribution lead the lowest distance, as shown in Table~\ref{tab:EMD}.

\begin{equation}
\begin{aligned}
&M_{ij} = \lVert {{dist}}_i - {{dist_{GT}}}_j \rVert_2, \\
&a_i    = \frac{1}{{{len(dist)}}}, \quad b_j = \frac{1}{{{len(dist_{GT})}}}, \\
&W({{dist}}, {{GT dist}}) = {\text{EMD}}(a, b, M).
\end{aligned}
\end{equation}

This result does not prove that $D_{SD}$ has multivariate normality, and the difference with the normal distribution is marginal. However, it is sufficient for our usage of the condition diffusion sampling for training.

\begin{table}[h]
\centering
\caption{Distance from GT embeddings.}
\small
\setcounter{table}{5}
\renewcommand{\arraystretch}{1.2} 
\label{tab:EMD}
\begin{tabular}{|l|c|}\hline 
\multicolumn{1}{|c|}{\textbf{Method}} & EMD     \\ \hline
Multivariate normal (Ours)                          & \textbf{ 244.22} \\ 
normal distribution for each axis                          &  244.31\\ 
uniform distribution                     &  1480.57\\ 
\hline 
\end{tabular}
\end{table}

\section*{E. Qualitative Results}
\label{sec:E}

\begin{figure*}[htb]
  \centering
  \includegraphics[width=1\linewidth]{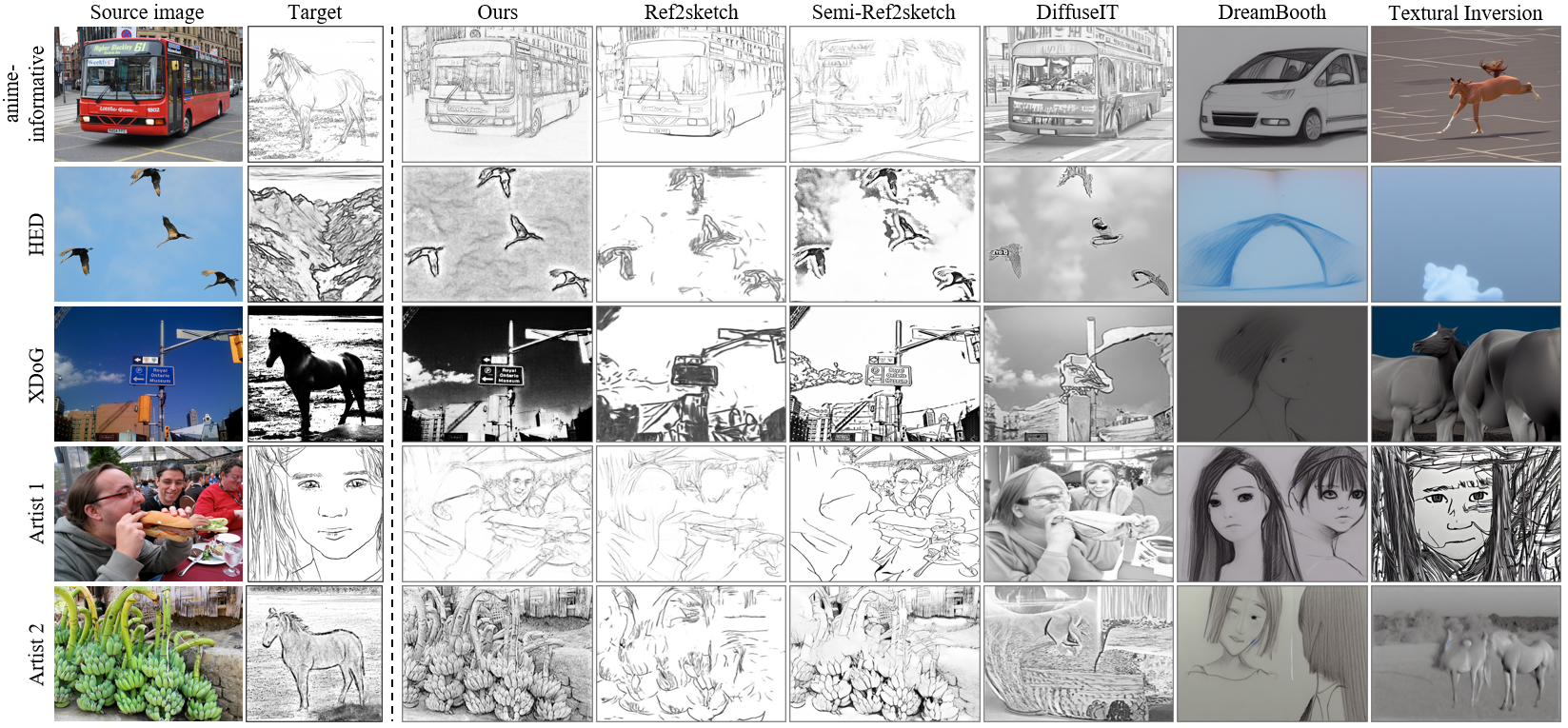}
  \caption{Qualitative comparison with alternative sketch extraction methods on COCO dataset.}
   \label{fig:comparsion_coco}
\end{figure*}

\begin{figure*}[htb]
  \centering
  \includegraphics[width=1\linewidth]{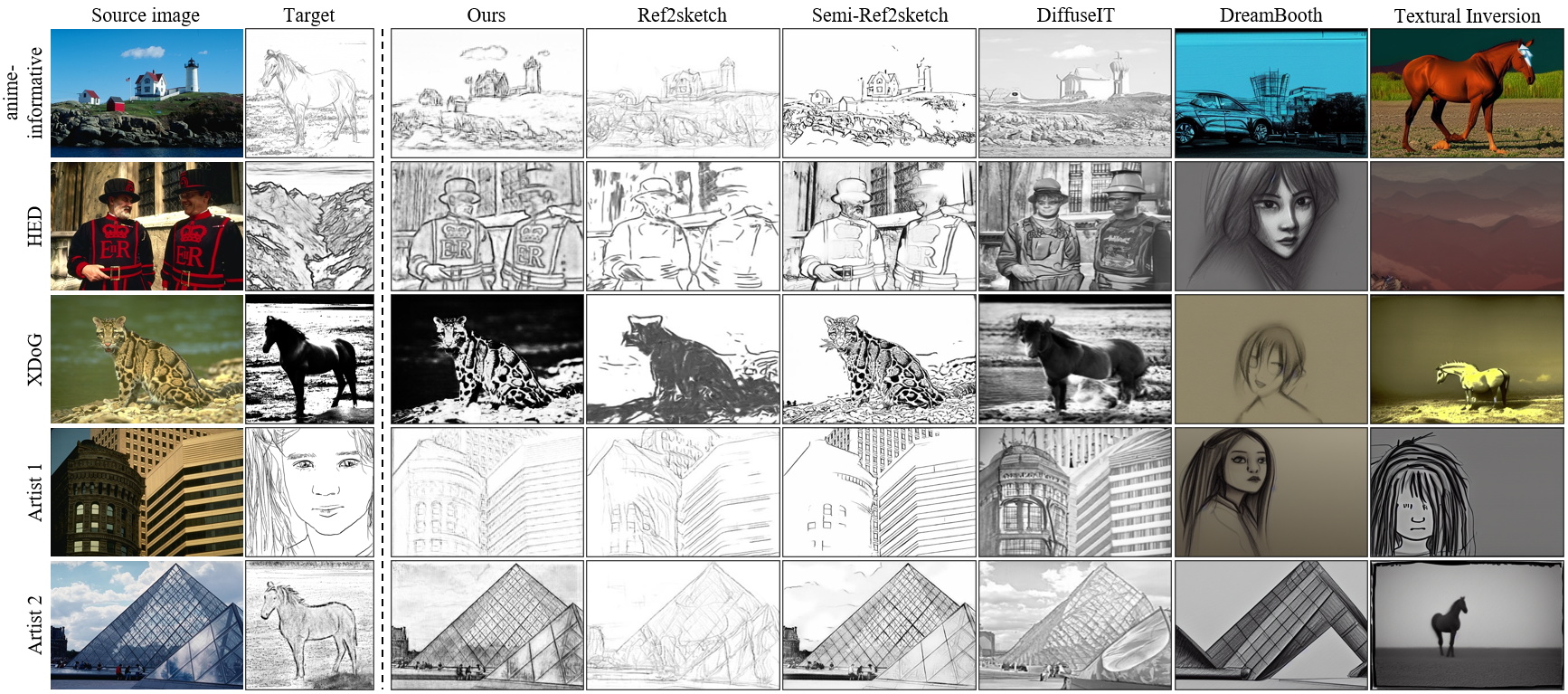}
  \caption{Qualitative comparison with alternative sketch extraction methods on BSDS500 dataset.}
   \label{fig:comparsion_bsds}
\end{figure*}

\begin{figure*}[htb]
  \centering
  \includegraphics[width=0.9\linewidth]{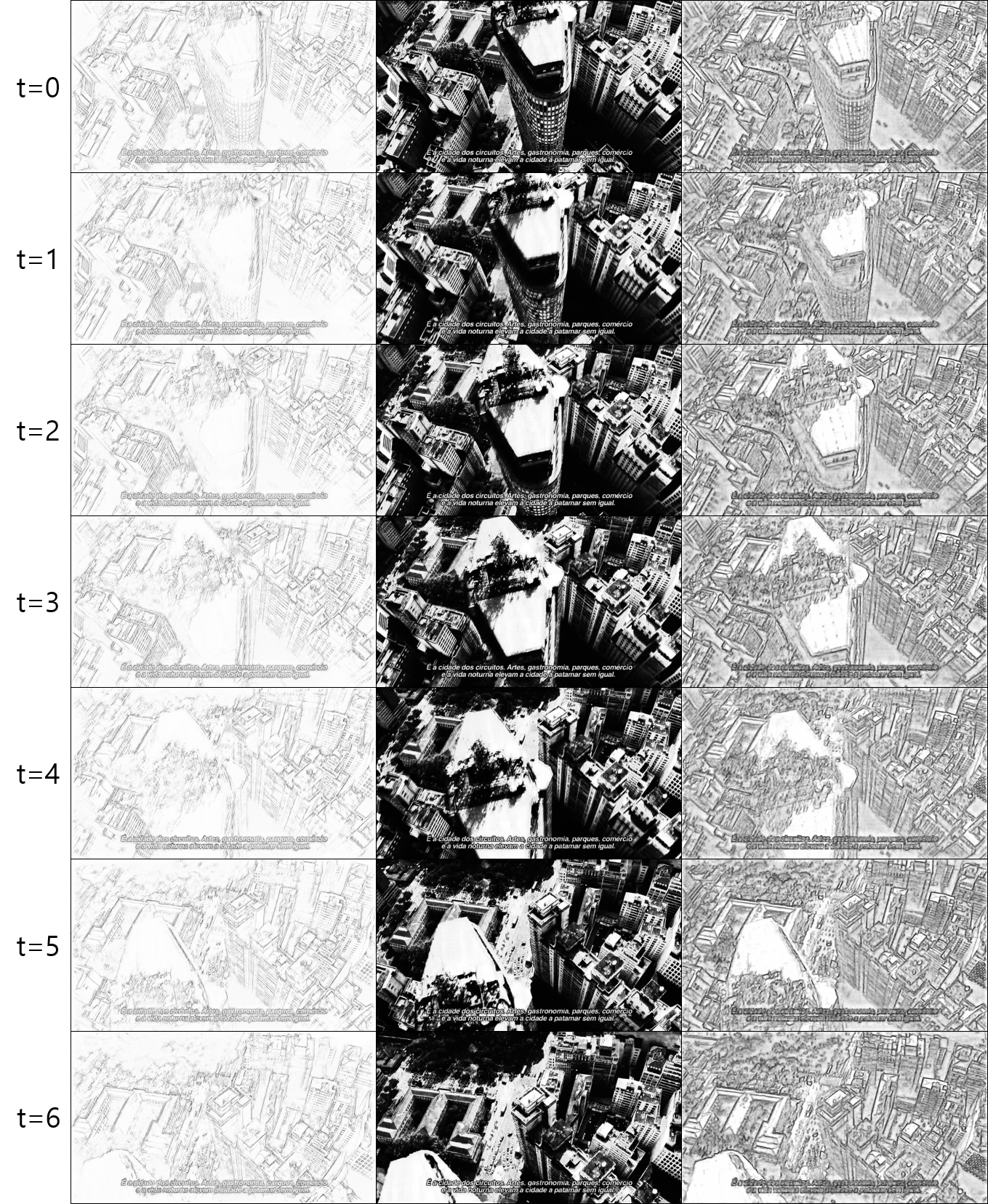}
  \caption{Qualitative examples of video sketch extraction.}
   \label{fig:video}
\end{figure*}

We present additional results from the baseline comparisons in Figure~\ref{fig:comparsion_coco} and ~\ref{fig:comparsion_bsds}. Each figure shows the results that compared $\text{DiffSketch}_{distilled}$ and the baseline methods on the COCO dataset~\cite{lin2015microsoft} and the BSDS500 dataset~\cite{martin2001database}, respectively. Addition to this, we also provide visual examples of video sketch extraction results on diverse domain including buildings, nature, and animals~\cite{vid1,vid2} using $\text{DiffSketch}_{distilled}$ in Figure~\ref{fig:video} and supplementary video.


\clearpage
\clearpage


\end{document}